\documentclass{article}
\usepackage{spconf,amsmath,graphicx}
\usepackage{booktabs}
\usepackage{multirow}
\usepackage{caption}
\usepackage[normalem]{ulem}

\graphicspath{{figures/}}

\title{VR-based generation of photorealistic synthetic data for training hand-object tracking models}
\name{Chengyan Zhang, Rahul Chaudhari}

\address{Technical University of Munich, 
        Chair of Media Technology, 
        Arcisstr. 21, 80333 Munich}

\begin{document}

\makeatletter
\let\@oldmaketitle\@maketitle
\renewcommand{\@maketitle}{\@oldmaketitle
  \includegraphics[width=\linewidth,height=14\baselineskip]
    {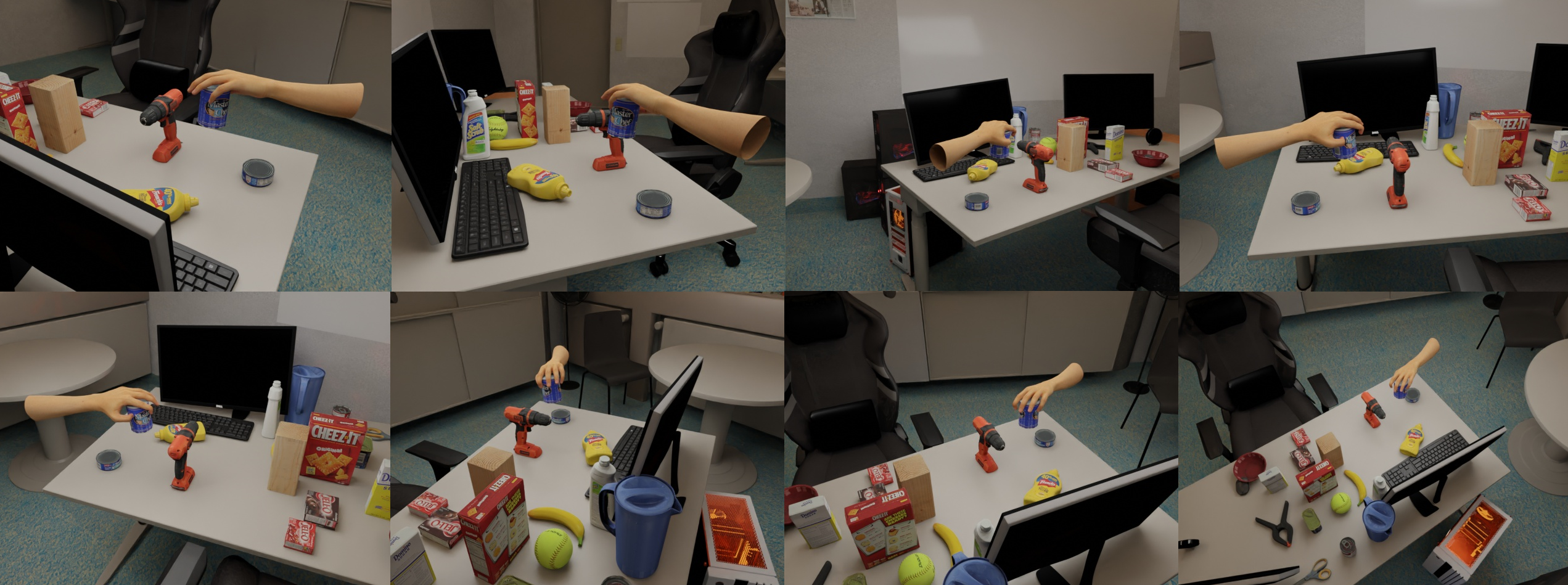}\bigskip}
\makeatother

\maketitle

\vfill
\begin{center}
    \footnotesize "This work has been submitted to the IEEE for possible publication. Copyright may be transferred without notice, after which this version may no longer be accessible."
    
\end{center}
\begin{abstract}
Supervised learning models for precise tracking of hand-object interactions (HOI) in 3D require large amounts of annotated data for training. Moreover, it is not intuitive for non-experts to label 3D ground truth (e.g. 6DoF object pose) on 2D images. To address these issues, we present "blender-hoisynth", an interactive synthetic data generator based on the Blender software. Blender-hoisynth can scalably generate and automatically annotate visual HOI training data. Other competing approaches usually generate synthetic HOI data compeletely without human input. While this may be beneficial in some scenarios, HOI applications inherently necessitate direct control over the HOIs as an expression of human intent. With blender-hoisynth, it is possible for users to interact with objects via virtual hands using standard Virtual Reality hardware. The synthetically generated data are characterized by a high degree of photorealism and contain visually plausible and physically realistic videos of hands grasping objects and moving them around in 3D. To demonstrate the efficacy of our data generation, we replace large parts of the training data in the well-known DexYCB dataset with hoisynth data and train a state-of-the-art HOI reconstruction model with it. We show that there is no significant degradation in the model performance despite the data replacement.
\end{abstract}


\section{Introduction}
\label{sec:intro}

The semantic understanding of hand-object interactions is dependent on low-level Computer vision tasks such as detection, localization, 6DoF pose estimation and tracking of both hands and objects. Solutions to these tasks, especially the ones based on Deep Learning, rely on good quality raw visual data and ground truth annotations. Several datasets have been introduced in the literature to address this need. Prominent examples include LineMOD~\cite{hinterstoisser2012model}, YCB-Video~\cite{xiang2018posecnn}, T-LESS~\cite{hodan2017t}, HOPE~\cite{tyree2022hope}, and HomebrewedDB~\cite{kaskman2019homebreweddb} for object-only data and DexYCB~\cite{DexYCB2021}, and H2O-3D~\cite{hampali2022keypointtransformer} for hand-object interaction data. A large amount of resources and effort are usually spent on building such datasets -- both for recording raw sensor data as well as for annotating the acquired data. Annotations such as marking 3D bounding box vertices on 2D images cannot be easily done by non-experts. Also, the dataset cannot be easily extended without investing a similar amount of effort as before.

To address the above issues, researchers have resorted to synthetically generated data. The popularity and efficacy of such approaches is driven by a steadily shrinking ``sim2real gap'' as more and more high quality 3D scans/models and renderers become available. Synthetic data generation based on virtual models is fast, inexpensive, and highly scalable. Furthermore, 2D and 3D data annotations such as object identities and bounding boxes, segmentation masks, 6D pose, etc. practically come for free with the raw data.  

\subsection{Contributions}
\label{subsec:contributions}
In this work we introduce a novel interactive synthetic data generator called ``blender-hoisynth", a short form for ``blender-based hand-object interaction synthetic data generator". Here, blender\footnote[1]{www.blender.org} refers to the free and open-source 3D software popularly used for creating object and scene models, animated films, visual effects, and interactive 3D applications. Blender-hoisynth allows intuitive interaction between virtual hands and objects in a scene using standard Virtual Reality (VR) hardware. The interaction can be saved as animation data, and subsequently re-rendered according to user requirements to produce synthetic raw sensor data and annotations useful for Machine Learning. An example of a multi-view render from hoisynth is shown in the picture on Page 1.

Secondly, we provide a synthetically generated hand-object interaction dataset with objects, grasping style and scenarios similar to the ones in the DexYCB dataset~\cite{DexYCB2021}. Note that the objects mentioned here are from the Yale-CMU-Berkeley (YCB) object set\footnote{https://registry.opendata.aws/ycb-benchmarks}. We name this dataset ``SynthDexYCB". This allows the study of performance on typical Computer Visions tasks while combining the synthetic data from blender-hoisynth with the original real-world DexYCB dataset in different ways. 

Thirdly, to show the efficacy of SynthDexYCB, we train a representative state-of-the-art hand-object mesh reconstruction model -- geometry-aligned Signed Distance Function (gSDF)~\cite{chen2023gsdf} -- on a mixture of synthetic and real data from DexYCB. We show that such a mixture does not significantly degrade in the model performance, indicating that the synthetic data quality is comparable to that of the real data.

\subsection{Game engine technology}
\label{subsec:intro_comparison_ge}
Most of the state-of-the-art synthetic data generators are based on commercial game engines such as Unity\footnote{www.unity.com} or Unreal Engine\footnote{www.unrealengine.com} (UE). To the best of our knowledge, blender-hoisynth is the first work relying on the free and open-source Blender game engine. Unlike Unity or UE, which only provide free versions for personal use, Blender can be used by teams of any size free of cost. Unlike Unity or UE, with blender-hoisynth creation of assets and HOI recordings can both happen in one place. We use a special fork of Blender called ``UPBGE"\footnote{www.upbge.org} after Uchronia Project Blender Game Engine. In the following, we use the names ``Blender" and ``UPBGE" interchangeably.

Works based on commercial game engines often only release the simulator containing the virtual world in binary form. Sometimes this may be due to licensing restrictions. But mostly the intention is to preclude the need for researchers to learn a new and complex game engine framework and programming language (e.g. C\# for Unity, C++ for UE). A Python API is additionally offered to interact with the simulator binary. However, this way the researcher only has an indirect and limited control over the 3D environment, making customization difficult. In the cases where the simulator source code is released, the code repository can easily run into tens of gigabytes in size. 

In contrast to the above approaches, blender-hoisynth is a single application purely written in Python, the most popular programming language in Computer Vision and AI at present. It is highly customizable, as any python library (e.g. a differentiable renderer) can be easily installed and directly used in the simulator. Python wrappers can be written for hardware drivers using SWIG\footnote{www.swig.org} enabling access to any VR hardware in blender-hoisynth. For robotics applications, it is possible to communicate with external hardware by using, e.g. the PySerial Python module or Python wrappers for the Robot Operating System (ROS\footnote{www.ros.org}) nodes. Due to all these features, we believe blender-hoisynth has the potential for widespread adoption among researchers.

\subsection{Applications}
\label{subsec:intro_applications}
Blender-hoisynth can export multi-view visual data and 2D/3D ground truth annotations of hand-object interactions. These data can be used by researchers for Computer Vision (CV) tasks such as 3D reconstruction, monocular or stereo depth estimation, semantic segmentation, normal estimation, etc. Also high-level CV tasks such as hand-object detection, pose estimation and tracking, visual object grasping and manipulation, etc. can benefit from training on the data exported from blender-hoisynth. 

Just as GPS on mobile phones enables several higher-level applications (navigation, neighborhood search), tracking of hands and objects can enable higher-level applications where humans manually interact with objects. Possible examples include: assistance/training in manual assembly of complex structures from parts, in medical surgery (visualization of target organs, prediction of complications), assistance for patients of dementia in conducting activities of daily living, etc.

\section{Related work}
\label{sec:related_work}

Synthetically generated data has been used increasingly in the past decade for training data-hungry Machine Learning models in Robotics and Computer Vision. Works in this direction include the Cornell House Agent Learning Environment (CHALET)~\cite{yan2018chalet}, Household Multimodal Environment (HoME)~\cite{brodeur2017home}, AI2-THOR~\cite{kolve2017ai2thor}, Multimodal Indoor Simulator (MINOS)~\cite{savva2017minos}, Gibson~\cite{xia2018gibson}, etc. Since these papers are not directly relevant for our work, we only mention their drawbacks here briefly: the lack of (photo-)realism, missing embodiment of the agent or its parts, inflexible camera configurations (number or poses of cameras), and only discrete interactions with the scene. They are generally characterized by a focus on larger building-scale 3D environments without necessarily a real-world correspondence, and on navigation rather than interaction. Another synthetic data generator called BlenderProc~\cite{Denninger2023} has been one of the main inspirations for our proposal. It is built on Blender and is characterized by photorealistic data and flexible camera configurations. However, it does not have an embodied agent and lacks real-time user interaction with the scene. 

We are interested in synthetic data generation from 3D scenes containing embodied hands interacting with objects through their hands. Works in this direction include: HOISIM~\cite{Zakour2021HOISIM}, VirtualHome~\cite{virtualhome2018}, Sims4Action~\cite{Sims4Action2021}, and ElderSIM~\cite{ElderSIM2023}. However, none of these works offer the user control over the virtual agent in real-time. Animations are either pre-recorded or generated automatically using standard game engine functionality such as navigation meshes, inverse kinematics for skeleton deformations, and collision detection between the agent and objects.

Finally, here we review some works allowing interactive control of the agent and 3D object grasping using VR hardware. The Modular Open Robots Simulation Engine (MORSE) presented in~\cite{echeverria2011morse} from 2011 was an early work in this direction. Similar to BlenderProc, MORSE is based on Blender and is one of the main inspirations for our work. Unfortunately, the project hasn't been actively developed for several years now. A very recent related approach, presented in~\cite{Leonardi2023}, combines virtual environments from the Habitat-Matterport 3D dataset and object-grasp pairs selected from DexGraspNet to generate fully-annotated HOI synthetic data. However, it does not have an embodied agent and lacks real-time user interaction with the scene.

Unreal Engine-based VRKitchen~\cite{VRKitchen} and Unity-based ThreeDWorld~\cite{gan2021threedworld} both support interaction with 3D objects using VR. Both these simulators follow the server-client architecture, with the server consisting of a binary of the game engine program and the client consisting of a Python program interacting with the server. As mentioned before in Section~\ref{subsec:intro_comparison_ge}, this way the user only has an indirect and limited control on the 3D environment, making customization difficult. 

Unreal Engine (UE)-based UnrealROX, presented by Martinez-Gonzalez et al. in~\cite{Martinez2019unrealrox}, comes closest to our work in terms of the goals as well as the approach. Among other things, like blender-hoisynth it can produce photorealistic data and ground truth annotations for user-recorded hand-object interactions. UnrealROX is a single application (no server-client) and the authors have made the entire code available. Since all this makes UnrealROX very promising, we experimented extensively with it before deciding to develop our own simulator. We found that UE has a complex architecture and workflows. UE applications must be programmed in C++ or the Blueprint system, both of which have a steep learning curve. The engine and the application together occupy 40-50GB of disk space and have high runtime resource requirements. UnrealROX was originally implemented using UE v4.18, which was released in 2017. The project has not been actively developed for more than two years at the time of this writing. We found it very difficult to upgrade UnrealROX to more recent versions of UE while maintaining original functionality. As an example, human skeleton tracking failed in going from UE v4.18 to v4.26. We believe that blender-hoisynth, with its reliance on the compact Blender Game Engine (couple of gigabytes in disk space and with moderate runtime requirements) and the Python programming language, can be more accessible to a wider audience both for use as well as for further customization. 

\section{blender-hoisynth}
\label{sec:blender_hoisynth}

\begin{figure*}[t!]
\centering
\includegraphics[width=0.9\linewidth]{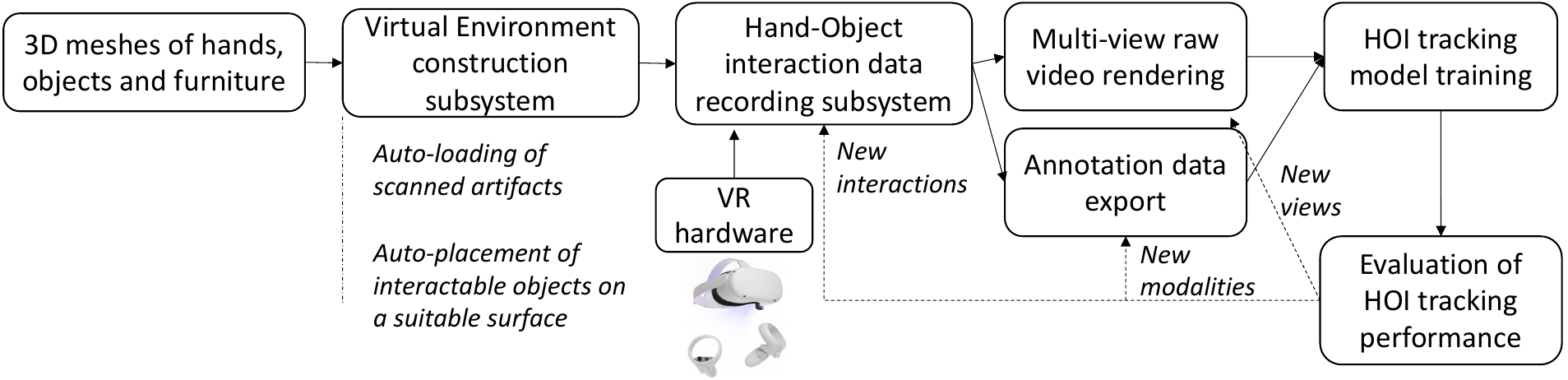}
\caption{Overview of blender-hoisynth and its usage.Fast iteration of HOI tracking model development can be achieved by customizing the data generator as required (dashed errors) in a feedback loop.}
\label{fig:block_schem_main}
\end{figure*}

The 3D environment in blender-hoisynth consists of textured 3D meshes of hands, objects, and the surrounding environment obtained from 3D scanners. Given the popularity of synthetic data generation methods in recent years, large repositories of high-quality object scans published by researchers can also be found and  imported. 

To interact with virtual objects, different VR hardware can be used for controlling the 6D pose of the virtual hand and the joint angles of the individual fingers. So far we have experimented with the Leap Motion Controller from Ultraleap Inc.\footnote{www.ultraleap.com}, the Meta Quest 2 VR controllers\footnote{www.meta.com/de/quest/products/quest-2/}, haptic I/O devices such as the Geomagic Touch\footnote{www.3dsystems.com/haptics-devices/touch}, and the OptiTrack\footnote{www.optitrack.com/} motion capture system.

Interactions typically consist of the hands approaching an object, grasping it, moving it around, and placing it somewhere. The hand-object interaction session can be recorded and played back later for review. If the user is satisfied with the created animation, the desired data can be rendered out. Figure~\ref{fig:block_schem_main} shows an overview of the main components of blender-hoisynth.

\subsubsection*{Hand-object interaction}
\label{subsubsec:hoi_subsys}

Blender-hoisynth firstly loads the 3D environment and hand-object meshes. Then it places the interactable objects on a surface chosen by the user with random poses. If desired, the user can also place the objects manually in the scene. Users immerse themselves into the 3D scene using a VR-headset and hand controllers. Presently we use the Meta Quest 2 headset for this purpose, since it is relatively inexpensive and is able to track the controllers in 3D space very precisely without the need for additional base stations. 

\begin{figure}[b!]
\centering
    \includegraphics[width=0.6\linewidth]{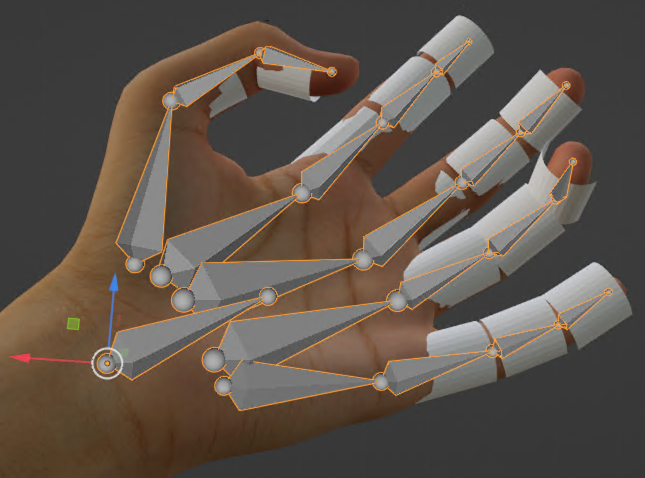}
    \caption{Virtual hand mesh showing the hierarchical bone rig and the half-cylinders collision sensors.}
    \label{fig:rigged_sensorized_hand}
\end{figure}

Hands are represented in the scene in the form of a mesh parented to a standard hierarchical bone rig as shown in Figure~\ref{fig:rigged_sensorized_hand}. Each bone in the rig is associated with vertices on the hand mesh with pre-defined weights that display realistic hand deformations. Controlling the relative rotations of the bones deforms the hand mesh accordingly. We drive the wrists with the 6D pose of the VR controllers. We use the analog trigger signal from the controller for driving the curling motion of the fingers when grasping an object. 

The fingers are clad from the inside with ``sensors" (visualized in Figure~\ref{fig:rigged_sensorized_hand}). These sensors are invisible to the user. When a collision is detected between the sensor and an object, the corresponding finger stops curling. When the thumb and at least one of the other four fingers are in contact with an object, the object is deemed to have been grasped. While being in the grasped state, the object is parented to the hand and thus follows the hand motion in 3D space. 

High-polygon object meshes in the scene are decimated to a couple of hundred polyons at most. At this level of decimation, the objects retain most of their characteristic shape making visually plausible grasping possible. These reductions are only necessary when the user is interacting with the scene in real-time. When the scene is rendered subsequently for data export, the original high-resolution meshes are used. During the interaction, poses of hands and objects are continuously being recorded and get saved as ``action" data in the Blender standard ``blend'' file format. 

\subsubsection*{Implementation in Blender/UPBGE}
\label{subsubsec:upbge_impl}
The blender game engine in the form of UPBGE (see Section~\ref{subsec:intro_comparison_ge}) enforces an object-oriented and modular architecture by default. Python components for hand/finger control, collision detection and response, object pose update, etc. are directly attached to the relevant simulation objects in the scene. They typically contain a function for initializing the object once at the beginning and other functions that are called at each simulation step. For example, when an object is deemed to be grasped by the hand, the object pose is updated at every frame to follow the hand. A single blend-file contains all the 3D scene data, the code, as well as the recorded interactions.

The desired data can be exported from the session recorded in blender-hoisynth in an offline rendering step. Depending on the camera viewpoints desired, the user can specify camera poses manually in the scene. It is also possible to only specify the number of cameras and let blender-hoisynth sample camera poses automatically, e.g. over a spherical surface. There is no restriction on the number of cameras that can be rendered. We set the virtual camera parameters to be the same as those of the real cameras in DexYCB.

\begin{figure*}[ht!]
    \centering
    \includegraphics[width=0.96\textwidth]{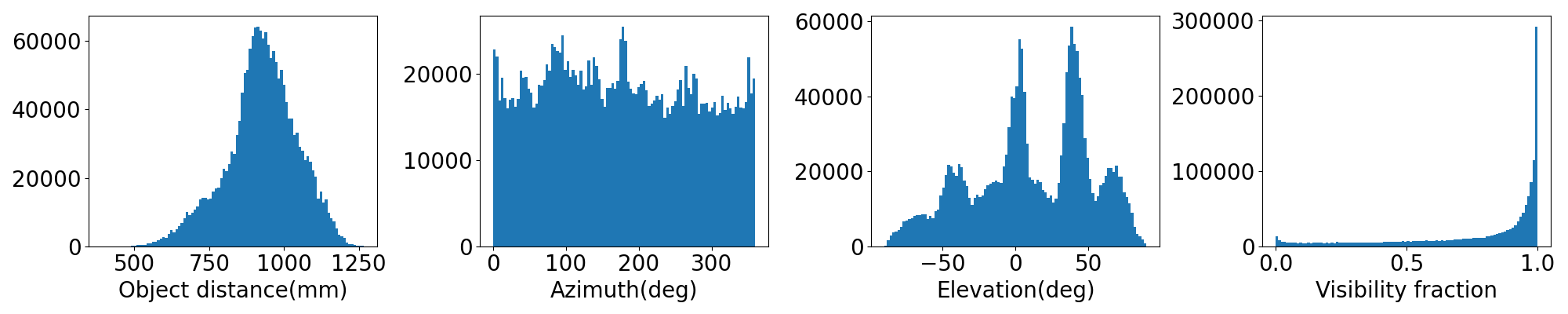}
    \vspace{-10pt}    
    \label{fig:s0_train_stitched}
\end{figure*}

\begin{figure*}[ht!]
    \centering
    \includegraphics[width=0.96\textwidth]{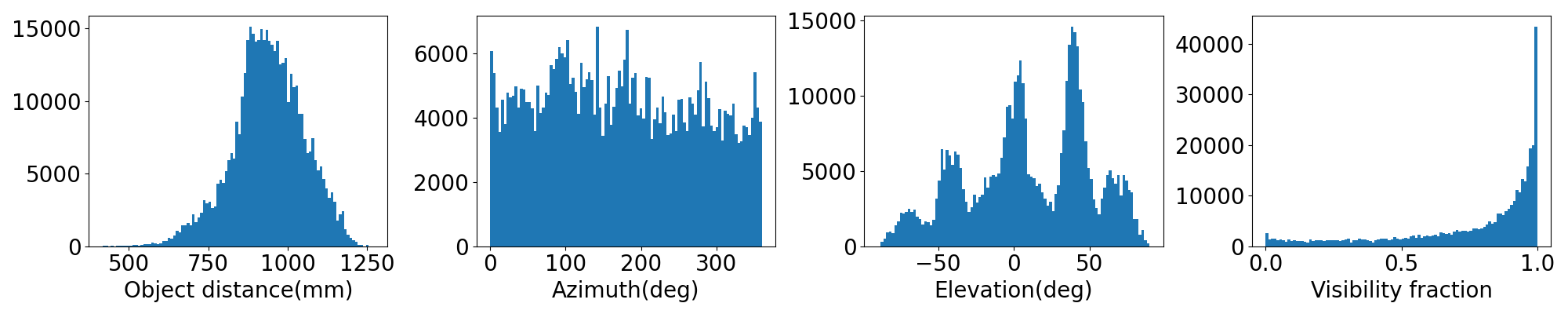}
    \vspace{-10pt}
    \caption{This figure shows that the ground truth parameter distributions of the synthetic data (bottom row) are qualitatively similar to those of the real data. Parameters shown are distance of the object from the camera optical center, object azimuths and elevations in the camera coordinate frame, and visibility fraction of the objects.}
    \label{fig:s0_train_stitched}
\end{figure*}

Once the session to be rendered and the cameras are configured, one of the in-built blender renderers (e.g. Cycles) can selected and the session can be rendered out in the form of multi-view RGB and depth image sequences. Furthermore, ground truth annotations such as object identities, bounding boxes and segmentation masks can be exported in the standard COCO format\footnote{https://cocodataset.org}, and 3D object poses can be exported in the BOP format\footnote{https://bop.felk.cvut.cz}. 21-DoF hand poses are exported in the MANO format\footnote{https://mano.is.tue.mpg.de/}. We rely on BlenderProc~\cite{Denninger2023} to provide the above export functionality, except for the hand poses for which we write a custom exporter.

\section{Dataset}
\label{sec:dataset}
Blender-hoisynth allows us to generate large-scale photorealistic visual data and annotations necessary for Machine Learning with very little effort. To investigate the quality of the data produced by blender-hoisynth, we generate a dataset analogous to the well-known DexYCB dataset from~\cite{DexYCB2021} and call it ``SynthDexYCB''.

The sequences in DexYCB consist of a subject reaching out to an object on the table, grasping it, lifting it up, before putting it down in roughly the same position again. SynthDexYCB captures interactions with 20 YCB objects from 8 views (same as DexYCB). It is very hard to extend the DexYCB dataset with the original interaction data for new camera views. In hoisynth, however, we can easily extend the number of views indefinitely by specifying new camera poses while retaining exactly the same interactions recorded by users before.

SynthDexYCB currently contains a total of 115,200 frames (right hands only) (as compared to 300,000 in DexYCB). This comes from ca. 200 hand-object interaction sequences, with 8 camera views rendered per sequence, and 72 frames per camera view. Human involvement was necessary only while recording these sequences in VR, which took ca. 7 hours. The Meta Quest 2 headset was used together with the "VR Scene Inspection" plugin for Blender. The authors closely replicated the grasps and motions in the original DexYCB videos in blender-hoisynth. While DexYCB acquired its ground truth annotations through a time-consuming and expensive crowd-sourcing process, the annotations in SynthDexYCB were generated automatically. Figure~\ref{fig:s0_train_stitched} shows a comparison of the distributions of ground truth parameters in the real data with those in the synthetic data.

\section{Experiments}
\label{sec:experiments}

In this section, we present SynthDexYCB-based training experiments for a representative 6D HOI tracking and reconstruction model -- gSDF~\cite{chen2023gsdf}. gSDF reconstructs the hand-object meshes in a HOI with a monocular RGB image as input. gSDF has been trained and evaluated on the DexYCB dataset in~\cite{chen2023gsdf}. We follow the same scheme and export data from hoisynth in the DexYCB format for training. We devise the following scenarios to evaluate the quality of the data in SynthDexYCB:

\textbf{Scenario s0 (synthetic data only)} investigates if the training converges at all with synthetic data only. Only 4 subjects are used. Training, validation, and test are all carried out on synthetic data only.

\textbf{Scenario s1 (real data only)} uses real data of 5 DexYCB subjects for training, validation, and test.

\textbf{Scenario s2 (real data only)} is the original s0 scenario with 10 subjects as defined in the DexYCB dataset and only uses real data for training, validation, and test.

\textbf{Scenario s3 (real + synthetic with replacement)} evaluates using synthetic and real data for training. We replace the data for subjects 1,2,3, and 10 in the original DexYCB dataset with synthetic data. While training is carried out on synthetic and real data, validation and test are carried out on real data.

\begin{table*}[h]
\centering
\caption{Hand-object reconstruction on gSDF~\cite{chen2023gsdf}. R = Real, S = Synth. Metrics shown ($_h$ = hand, $_o$ = object): $CD$ = chamfer distance in $cm^2$, $FS@X$ = F-score at threshold $X mm$, $E_h$ = mean hand joint error in $cm$, $E_h$ = mean object error in $cm$.}
\small
\renewcommand{\arraystretch}{1.0}
\begin{tabular}{@{}ccccccccccc@{}}
\toprule
Scenario & \begin{tabular}[c]{@{}c@{}}Synthetic \\ Subjects\end{tabular} & $CD_{h} \downarrow$ & $FS_{h}@1\uparrow$ & $FS_{h}@5\uparrow$ & $E_{h} \downarrow$& $CD_{o} \downarrow$ & $FS_{o}@5\uparrow$ & $FS_{o}@10\uparrow$ & $E_{o} \text{(center)} \downarrow$ & $E_{o} \text{(corner)} \downarrow$ \\
\midrule
s0 (S) &1,2,3,10&\textbf{0.260}&\textbf{0.173}&\textbf{0.814}&24.188&2.110&0.407&0.661&20.994&100.596\\
\hline
s1 (R)
&None&0.301&0.172&0.800&19.864&2.338&0.393&0.655&19.336&96.258\\
\hline
s2 (R)
&None&0.302&0.172&0.800&\textbf{19.325}&2.024&0.407&0.670&\textbf{19.319}&77.783\\
\hline
\multirow{6}{*}{s3 (R+S)}&10&0.310&0.170&0.796&19.733&2.056&0.405&0.667&19.546&77.486\\ 
    &1,10&0.316&0.170&0.795&19.457&1.929&0.409&0.673&19.773&\textbf{72.646}\\ 
    &1,2,10&0.331&0.167&0.789&19.643&1.960&0.410&0.671&19.770&74.501\\ 
    &1,3,10&0.337&0.164&0.783&20.260&1.934&0.406&0.670&19.840&74.531\\ 
    &2,3,10&0.330&0.165&0.784&19.837&\textbf{1.884}&\textbf{0.413}&\textbf{0.674}&19.935&73.916\\ 
    &1,2,3,10&0.344&0.164&0.782&19.915&1.955&0.407&0.669&19.847&73.905\\ 

\bottomrule
\end{tabular}
\label{tab:results}
\end{table*}

\section{Results}
\label{sec:results}
The hand-object mesh reconstruction performance on gSDF is shown in Table~\ref{tab:results}. Scenarios s0 and s1 shows outlier values for $E_o(corner)$, since they use fewer subjects than s2 and s3. Scenario s0 has the best performance for hand mesh reconstruction due to highly accurate ground truth poses for hands in the synthetic data. 

Comparing s2 and s3, the variance along columns is quite small. From this, we conclude that replacing parts of the real data with synthetic data does not deteriorate performance significantly. In other words, the generated synthetic data is qualitatively comparable to the real data. 

The performance metrics for hand mesh reconstruction are slightly better for the purely real data than for the real-synthetic combination. There are a couple of reasons for this. Currently, it is impossible to control fingers individually in hoisynth, which may lead to artifacts not present in the real data. Secondly, while 3D scans for YCB objects are available, DexYCB does not provide 3D scans of the subjects' hand meshes. Hence, we were forced to use 3rd party hand meshes, which are different in shape and appearance from the DexYCB subjects' hands. 

The real-synthetic combination provides slightly better results than pure real data for object reconstruction. A possible reason for this is that the accuracy of ground truth in the synthetic data is better than that of the manually labeled ground truth provided by DexYCB.

\section{Conclusion}
\label{sec:conclusion}
In this work, we have presented a novel interactive synthetic HOI data generator called ``blender-hoisynth'', which is characterized by a high degree of realism (both visual as well as physical). Unlike other related works, it enables VR-based user interaction with virtual objects. Thus training data is generated with the human in the loop, which is especially important for such a human-centric topic as hand-object tracking and reconstruction. Furthermore, we show that there is no significant degradation in HOI tracking model performance after replacing parts of the DexYCB dataset with synthetic data.

We plan to support bimanual HOIs in the future. The curling of fingers during grasping is currently driven by a single analog button on the VR controller. Although the timing and speed of the grasp can be controlled by the user, and the bone trajectories and the bone trjactories can be reasonably randomized, individual finger control is not yet possible. In the future, we plan to support full finger control based on visual hand tracking sensors. The deformations of the hand mesh can be made much more realistic and personalized by relying on works such as DeepHandMesh~\cite{Moon_2020_ECCV_DeepHandMesh}. On the user interaction side, the intuitiveness of interactions in blender-hoisynth can benefit immensely from the inclusion of haptic feedback. 

Finally, the sim2real gap can be further reduced by using Image-to-Image translation approaches trained on synthetic and real data of the same scene. Replacing the in-built physics based renderers in blender by state-of-the-art differentiable renderers could also be an interesting avenue to explore.

\bibliographystyle{IEEEbib}
\bibliography{2024-icip-blender-hoisynth}

\begin{thebibliography}{10}

\bibitem{hinterstoisser2012model}
S.~Hinterstoisser, V.~Lepetit, S.~Ilic, S.~Holzer, G.~Bradski, K.~Konolige, and
  N.~Navab,
\newblock ``Model based training, detection and pose estimation of texture-less
  3d objects in heavily cluttered scenes,''
\newblock {\em International Journal of Computer Vision}, vol. 97, no. 2, pp.
  238--252, 2012.

\bibitem{xiang2018posecnn}
Y.~Xiang, T.~Schmidt, V.~Narayanan, and D.~Fox,
\newblock ``Posecnn: A convolutional neural network for 6d object pose
  estimation in cluttered scenes,''
\newblock 2018.

\bibitem{hodan2017t}
T.~Hodan, P.~Haluza, Š. Obdržálek, and J.~Matas,
\newblock ``T-less: An rgb-d dataset for 6d pose estimation of texture-less
  objects,''
\newblock {\em IEEE Transactions on Pattern Analysis and Machine Intelligence
  (TPAMI)}, vol. 41, no. 12, pp. 2990--3004, 2019.

\bibitem{tyree2022hope}
S.~Tyree, J.~Tremblay, T.~To, J.~Cheng, T.~Mosier, J.~Smith, and S.~Birchfield,
\newblock ``6-dof pose estimation of household objects for robotic
  manipulation: An accessible dataset and benchmark,''
\newblock in {\em International Conference on Intelligent Robots and Systems
  (IROS)}, 2022.

\bibitem{kaskman2019homebreweddb}
R.~Kaskman, S.~Zakharov, I.~Shugurov, and S.~Ilic,
\newblock ``Homebreweddb: Rgb-d dataset for 6d pose estimation of 3d objects,''
\newblock {\em International Conference on Computer Vision (ICCV) Workshops},
  2019.

\bibitem{DexYCB2021}
Y.-W. Chao, W.~Yang, Y.~Xiang, P.~Molchanov, A.~Handa, J.~Tremblay, Y.~S.
  Narang, K.~{Van Wyk}, U.~Iqbal, S.~Birchfield, J.~Kautz, and D.~Fox,
\newblock ``{DexYCB}: A benchmark for capturing hand grasping of objects,''
\newblock in {\em IEEE/CVF Conference on Computer Vision and Pattern
  Recognition (CVPR)}, 2021.

\bibitem{hampali2022keypointtransformer}
S.~Hampali, S.~D. Sarkar, M.~Rad, and V.~Lepetit,
\newblock ``Keypoint transformer: Solving joint identification in challenging
  hands and object interactions for accurate 3d pose estimation,''
\newblock in {\em CVPR}, 2022.

\bibitem{chen2023gsdf}
Z.~Chen, S.~Chen, C.~Schmid, and I.~Laptev,
\newblock ``{gSDF}: {Geometry-Driven} signed distance functions for {3D}
  hand-object reconstruction,''
\newblock in {\em IEEE/CVF Conference on Computer Vision and Pattern
  Recognition (CVPR)}, 2023.

\bibitem{yan2018chalet}
C.~Yan, D.~Misra, A.~Bennett, A.~Walsman, Y.~Bisk, and Y.~Artzi,
\newblock ``Chalet: Cornell house agent learning environment,''
\newblock in {\em Proceedings of the AAAI Conference on Artificial Intelligence
  (AAAI)}, 2018.

\bibitem{brodeur2017home}
S.~Brodeur, E.~Perez, A.~Anand, F.~Golemo, L.~Celotti, F.~Strub, J.~Rouat,
  H.~Larochelle, and A.~Courville,
\newblock ``Home: a household multimodal environment,'' 2017.

\bibitem{kolve2017ai2thor}
E.~Kolve, R.~Mottaghi, D.~Gordon, Y.~Zhu, A.~Gupta, and A.~Farhadi,
\newblock ``Ai2-thor: An interactive 3d environment for visual ai,''
\newblock in {\em Proceedings of the IEEE Conference on Computer Vision and
  Pattern Recognition (CVPR)}, 2017, pp. 6195--6204.

\bibitem{savva2017minos}
M.~Savva, A.~X. Chang, A.~Dosovitskiy, T.~Funkhouser, and V.~Koltun,
\newblock ``Minos: Multimodal indoor simulator for navigation in complex
  environments,''
\newblock in {\em Proceedings of the IEEE Conference on Computer Vision and
  Pattern Recognition (CVPR)}, 2017, pp. 6402--6410.

\bibitem{xia2018gibson}
F.~Xia, A.~R. Zamir, Z.-Y. He, A.~Sax, J.~Malik, and S.~Savarese,
\newblock ``Gibson env: Real-world perception for embodied agents,''
\newblock in {\em Proceedings of the IEEE Computer Vision and Pattern
  Recognition (CVPR)}, 2018.

\bibitem{Denninger2023}
M.~Denninger, D.~Winkelbauer, M.~Sundermeyer, W.~Boerdijk, M.~Knauer, K.~H.
  Strobl, M.~Humt, and R.~Triebel,
\newblock ``Blenderproc2: A procedural pipeline for photorealistic rendering,''
\newblock {\em Journal of Open Source Software}, vol. 8, no. 82, pp. 4901,
  2023.

\bibitem{Zakour2021HOISIM}
M.~Zakour, A.~Mellouli, and R.~Chaudhari,
\newblock ``Hoisim: Synthesizing realistic 3d human-object interaction data for
  human activity recognition,''
\newblock in {\em 2021 30th IEEE International Conference on Robot \& Human
  Interactive Communication (RO-MAN)}, 2021, pp. 1124--1131.

\bibitem{virtualhome2018}
X.~Puig, J.~Ra, M.~Boben, T.~Gangwani, and A.~Torralba,
\newblock ``Virtualhome: Simulating household activities via programs,''
\newblock in {\em 2018 IEEE Conference on Computer Vision and Pattern
  Recognition (CVPR)}, 2018, pp. 8494--8503.

\bibitem{Sims4Action2021}
A.~Roitberg, D.~Schneider, A.~Djamal, C.~Seibold, S.~Reiß, and
  R.~Stiefelhagen,
\newblock ``Let\'s play for action: Recognizing activities of daily living by
  learning from life simulation video games,''
\newblock in {\em 2021 IEEE/RSJ International Conference on Intelligent Robots
  and Systems (IROS)}, 2021, pp. 8563--8569.

\bibitem{ElderSIM2023}
H.~Hwang, C.~Jang, G.~Park, J.~Cho, and I.-J. Kim,
\newblock ``Eldersim: A synthetic data generation platform for human action
  recognition in eldercare applications,''
\newblock {\em IEEE Access}, vol. 11, pp. 9279--9294, 2023.

\bibitem{echeverria2011morse}
G.~Echeverria, N.~Lassabe, Arnaud Degroote, and Séverin Lemaignan,
\newblock ``Modular open robots simulation engine: Morse,''
\newblock in {\em Proceedings of the IEEE/RSJ International Conference on
  Intelligent Robots and Systems (IROS)}, 2011, pp. 1742--1748.

\bibitem{Leonardi2023}
R.~Leonardi, A.~Furnari, F.~Ragusa, and G.~M. Farinella,
\newblock ``Are synthetic data useful for egocentric hand-object interaction
  detection? an investigation and the hoi-synth domain adaptation benchmark,''
\newblock {\em CoRR}, vol. abs/2312.02672, 2023.

\bibitem{VRKitchen}
X.~Gao, R.~Gong, T.~Shu, X.~Xie, S.~Wang, and S.-C. Zhu,
\newblock ``Vrkitchen: an interactive 3d virtual environment for task-oriented
  learning,''
\newblock {\em arXiv}, vol. abs/1903.05757, 2019.

\bibitem{gan2021threedworld}
C.~Gan, J.~Schwartz, S.~Alter, D.~Mrowca, M.~Schrimpf, J.~Traer, J.~De Freitas,
  J.~Kubilius, A.~Bhandwaldar, N.~Haber, M.~Sano, K.~Kim, E.~Wang,
  M.~Lingelbach, A.~Curtis, K.~Feigelis, D.~M. Bear, D.~Gutfreund, D.~Cox,
  A.~Torralba, J.~J. DiCarlo, J.~B. Tenenbaum, J.~H. McDermott, and D.~L.~K.
  Yamins,
\newblock ``Threedworld: A platform for interactive multi-modal physical
  simulation,'' 2021.

\bibitem{Martinez2019unrealrox}
P.~Martinez-Gonzalez, S.~Oprea, A.~Garcia-Garcia, A.~Jover-Alvarez,
  S.~Orts-Escolano, and J.~Garcia-Rodriguez,
\newblock ``{UnrealROX}: An extremely photorealistic virtual reality
  environment for robotics simulations and synthetic data generation,''
\newblock {\em Virtual Reality}, 2019.

\bibitem{Moon_2020_ECCV_DeepHandMesh}
G.~Moon, T.~Shiratori, and K.~M. Lee,
\newblock ``Deephandmesh: A weakly-supervised deep encoder-decoder framework
  for high-fidelity hand mesh modeling,''
\newblock in {\em European Conference on Computer Vision (ECCV)}, 2020.

\end{thebibliography}

\end{document}